\documentclass[10pt,twocolumn,letterpaper]{article}

\usepackage{cvpr}
\usepackage{times}
\usepackage{epsfig}
\usepackage{graphicx}
\usepackage{amsmath}
\usepackage{amssymb}
\usepackage{array}
\usepackage{booktabs}
\usepackage{mathrsfs}
\usepackage{algorithm}
\usepackage{algorithmic}
\usepackage{multirow}
\usepackage{mathtools}
\usepackage{enumitem}
\usepackage[normalem]{ulem}
\usepackage{color}
\usepackage{caption}
\usepackage{bbm}
\usepackage{setspace}

\newcommand{\MYWHILE}{\item[\mywhile]}
\newcommand{\mywhile}{\textbf{repeat}}

\newcommand{\MYENDWHILE}{\item[\myendwhile]}
\newcommand{\myendwhile}{\textbf{until}}

\usepackage[pagebackref=true,breaklinks=true,letterpaper=true,colorlinks,bookmarks=false]{hyperref}

\cvprfinalcopy % *** Uncomment this line for the final submission

\ifcvprfinal\fi
\begin{document}

%%%%%%%%% TITLE
\title{Actions $\sim$ Transformations}

\author{Xiaolong Wang$^{1}$\thanks{Work was  done while Xiaolong Wang was an intern at AI2.}  \quad     \quad  Ali Farhadi$^{2,3}$    \quad  \quad  Abhinav Gupta$^{1,3}$  \\
$^1$Carnegie Mellon University \quad $^2$University of Washington \quad $^3$The Allen Institute for AI\\
%{\tt\small \{xiaolonw, abhinavg\}@cs.cmu.edu}
}

\maketitle
\vspace{-0.2in}
%%%%%%%%% ABSTRACT
\begin{abstract}
\vspace{-0.1in}
What defines an action like ``kicking ball''? We argue that the true meaning of an action lies in the change or transformation an action brings to the environment. In this paper, we propose a novel representation for actions by modeling an action as a transformation which changes the state of the environment before the action happens (precondition) to the state after the action (effect). Motivated by recent advancements of video representation using deep learning, we design a Siamese network which models the action as a transformation on a high-level feature space. We show that our model gives improvements on standard action recognition datasets including UCF101 and HMDB51. More importantly, our approach is able to  generalize beyond learned action categories and shows significant performance improvement on cross-category generalization on our new ACT dataset.
\end{abstract}

%%%%%%%%% BODY TEXT
\vspace{-3mm}
\section{Introduction}
\begin{figure}
\includegraphics[width=0.5\textwidth]{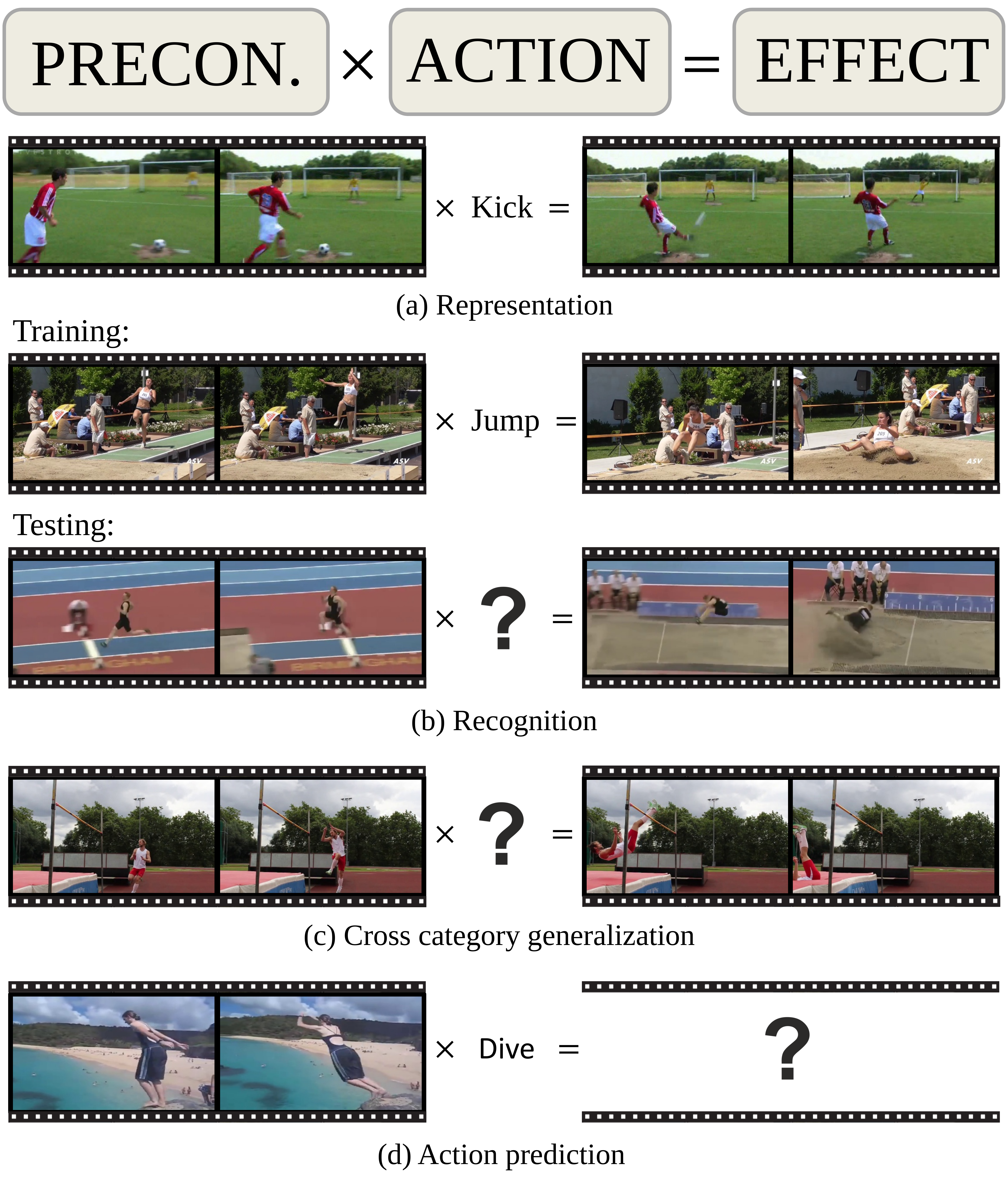}
\vspace{-0.2in}
\caption{\label{fig:teaser} \footnotesize{
We represent actions as the transformations from precondition to effect. (a) For example, the precondition of kicking is the player running towards the ball and the effect is the ball flies away. By using this representation, we can (b) perform action recognition given the training data on long jump, and (c) generalize the classifier to high jump in testing. (d) Moreover,  we can perform visual prediction given the precondition.} }
\vspace{-0.2in}
\end{figure}

\vspace{-1mm}
Consider the  ``soccer kicking'' action shown in Figure~\ref{fig:teaser}. What is the right representation for the recognition of such an action? Traditionally, most research in action recognition has focused on learning discriminative classifiers on hand-designed features such as HOG3D~\cite{HOG3D} and IDT~\cite{WangIDT13}. Recently, with the success of deep learning approaches, the focus has moved from hand-designed features to building end-to-end learning systems. However, the basic philosophy remains the same: representing action implies encoding the appearance and motion of the actor. But are actions all about appearance and motion? 

We argue that the true essence of an action lies in the change or the transformation an action brings to the environment and most
often these changes can be encoded visually. For example, the essence of ``soccer kicking'' lies in the state change of the ball (acceleration) caused by the leg of the player. What if we try to represent actions based on these changes rather than appearance and motion?

In this paper, we propose representing actions as transformations in the visual world. We argue that current action recognition approaches tend to overfit by focusing on scene context and hence do not generalize well. This is partly because of the lack of diversity in action recognition datasets compared to object recognition counterparts. In this paper, we overcome this problem by forcing a representation to explicitly encode the change in the environment: the inherent reason that convinced the agent to perform the action. Specifically, each action is represented as a transformation that changes the state of the environment from what it was before the action to what  it will be after it. Borrowing the terminology from NLP, we refer to the state before the action as the precondition state and the state after the action as the effect state, as Fig.~\ref{fig:teaser} illustrates.  We build a discriminative model of transformation by using a Siamese network architecture (similar to ~\cite{Chopra05,Bromley93}) where the action is represented as a linear transformation between the final fully connected layers of the two towers representing the precondition and effect states of an action.

Our experimental evaluations show that our representation is well suited for classical action recognition and gives state of the art results on standard action recognition dataset such as UCF101~\cite{UCF12}. However, in order to test the robustness of our representation, we also test our model for cross-category generalization.
While overfitted representations would perform competitively on current action recognition datasets (due to lack of diversity), the true test lies in their ability to generalize beyond learned action categories. For example, how would a model learned on ``opening a window'' generalize to recognize ``opening the trunk of the car''? How about generalizing from a model trained on climbing a cliff to recognize climbing a tree? Our experimental evaluations show that our representation allows successful transfer of models across action categories (Fig.~\ref{fig:teaser} (c)). Finally, our transformation model can also be used to predict what is about to happen (Fig.~\ref{fig:teaser} (d)).

Our contributions include: (a) a new representation of actions based on transformations in visual world; (b) state of the art performance on an existing action recognition dataset: UCF101; (c) addressing the cross-category generalization task and proposing a new dataset, ACT, consisting of 43 categories of actions, which can be further grouped to 16 classes, and 11234 videos; (d) results on prediction task for our ACT dataset. 

\vspace{-2mm}
\section{Related Work}
\vspace{-2mm}

Action recognition has been extensively studied in computer vision. Lack of space does not allow a comprehensive literature review (see \cite{poppe2010survey} for a survey).

{\bf Hand-crafted representations} have been conventionally  used to describe patches centered at Space Time Interest Points (STIP)~\cite{STIP05}. Most successful examples are 3D Histogram of Gradient (HOG3D)~\cite{HOG3D}, Histogram of Optical Flow (HOF)~\cite{HOF}, and Motion Boundary Histogram~\cite{MBH06}. Mid- to high-level representation are also used to model complex actions ~\cite{YaleS13,YangGreg11,Corso12,Zhuowen13,ArpitJ13,lan2015iccv}. More recently, trajectory based approaches~\cite{WangIDT13,iDTHSV14,Matikainen09,Wang11,Jiang12,Peng14,Lan15} have shown significant improvement in action recognition.

{\bf Learned representations} with deep learning have recently produced state of the art results in action recognition~\cite{3DCNN,Karpathy14,2stream14,TDD15,Taylor10,Tran15,Le11,Georgia15,Xu15}. Karpathy et al.~\cite{Karpathy14} proposed to train Convolutional Neural Networks (ConvNets) for video classification on the  Sports-1M dataset. To better capture motion information in video, Simonyan et al.~\cite{2stream14} introduced a Two Stream framework to train two separate ConvNets for motion and color. Based on this work, Wang et al.~\cite{TDD15} extracted deep feature and conducted trajectory constrained pooling to aggregate convolutional feature as video representations. In our paper, we also train the networks taking RGB frames and optical flows as inputs. 

%Instead of learning features with 2D CNNs, Tran et al.~\cite{Tran15} proposed to learn 3D CNNs, which offers effective and compact deep features. In this paper, 

{\bf Temporal structure} of videos have also been shown effective in action recognition~\cite{KevinT12,Basura15,Sun13,Izadinia12,Rohrbach12}. For example, Tang et al.~\cite{KevinT12} proposed an HMM model to model the duration as well as the transitions of states in event video. Fernando et al.~\cite{Basura15} learned ranking functions for each video and tried to capture video-wide temporal information for action recognition. Recurrent Neural Networks have also been used to encode temporal information for learning video representations~\cite{Srivastava15,Donahue15,LSTMgoogle15,Sun15,Wu15,Sun_2015_ICCV}. Srivastava et al.~\cite{Srivastava15} proposed to learn video representations with LSTM in an unsupervised manner. Ng et al.~\cite{LSTMgoogle15} proposed to extract features with a Two Stream framework and perform LSTM fusion for action recognition. However, these HMM, RNN and recent LSTM approaches model a sequence of transformation across frames or key frames; whereas in our framework we model action as a transformation between precondition and effect of action. Note that the location of these frames are latent in our model.

% Note that our approach also applies a constraint such that precondition and effect frames for an action lie close-by the embedding space.

The most similar work to ours is from Fathi et  al.~\cite{Fathi13} where the change in the state of objects are modeled using hand-crafted features in ego-centric videos of 7 activities. We differ from~\cite{Fathi13} in that we learn representations which enable explicit encoding of actions as transformations. Our representations not only produce state of the art generic action recognition results, but also allow predictions of the outcome of actions as well as cross-category model generalization. To model the transformation, we apply a Siamese network architecture in this paper, which is also related to the literature using deep metric learning~\cite{Hoffer14,Hadsell06,Edgar14,Hu_2014_CVPR,WangUnsup15,Dinesh15}.

\vspace{-2mm}
\section{Dataset}
\vspace{-2mm}

To study action recognition, several datasets have been compiled. Early datasets (e.g. Weizmann, KTH, Hollywood2, UCF Sports, UCF50) are too small for training ConvNets. Recently, a few large-scale video datasets have been introduced (e.g. CCV~\cite{icmr11:consumervideo}, Sports-1M~\cite{Karpathy14}, ActivityNet~\cite{Heilbron15},  THUMOS~\cite{Jiang14} and FGA-240 datasets~\cite{Sun15}). Unfortunately, some of these datasets have untrimmed videos without localization information for short term actions. The most commonly studied datasets are UCF101~\cite{UCF12} and HMDB51~\cite{HMDB11}. The UCF101 dataset lacks the desired diversity among   videos in each class. The HMDB51 dataset does not have enough videos compared to UCF101, and some of the videos are hard to recognize. But more importantly, none of these datasets is suitable for our task of examining cross category generalization of actions.

In this paper, we argue for cross-category generalization as a litmus test for action recognition. We believe that the cross-category recognition task should not allow approaches to overfit to action classes based on contextual information. For this task, we propose a dataset, namely ACT dataset. In this dataset, we collected 11234 video clips with 43 classes. These 43 classes can be further grouped into 16 super-classes. For example, we have classes such as kicking bag and kicking people, they all belong to the super-class kicking; swinging baseball, swinging golf and swinging tennis can be grouped into swinging. Thus, the categories are arranged in a 2-layer hierarchy. The higher layer represents super-classes of actions such as kicking and  swinging. Each super-class has different sub-categories which are the same action under different subjects, objects and scenes. During the dataset collection, we also ensured that we only consider high resolution and diverse videos.

%; climbing-rock, climbing-rope and climbing-tree can be grouped to climbing.

\begin{figure}
\includegraphics[width=0.45\textwidth]{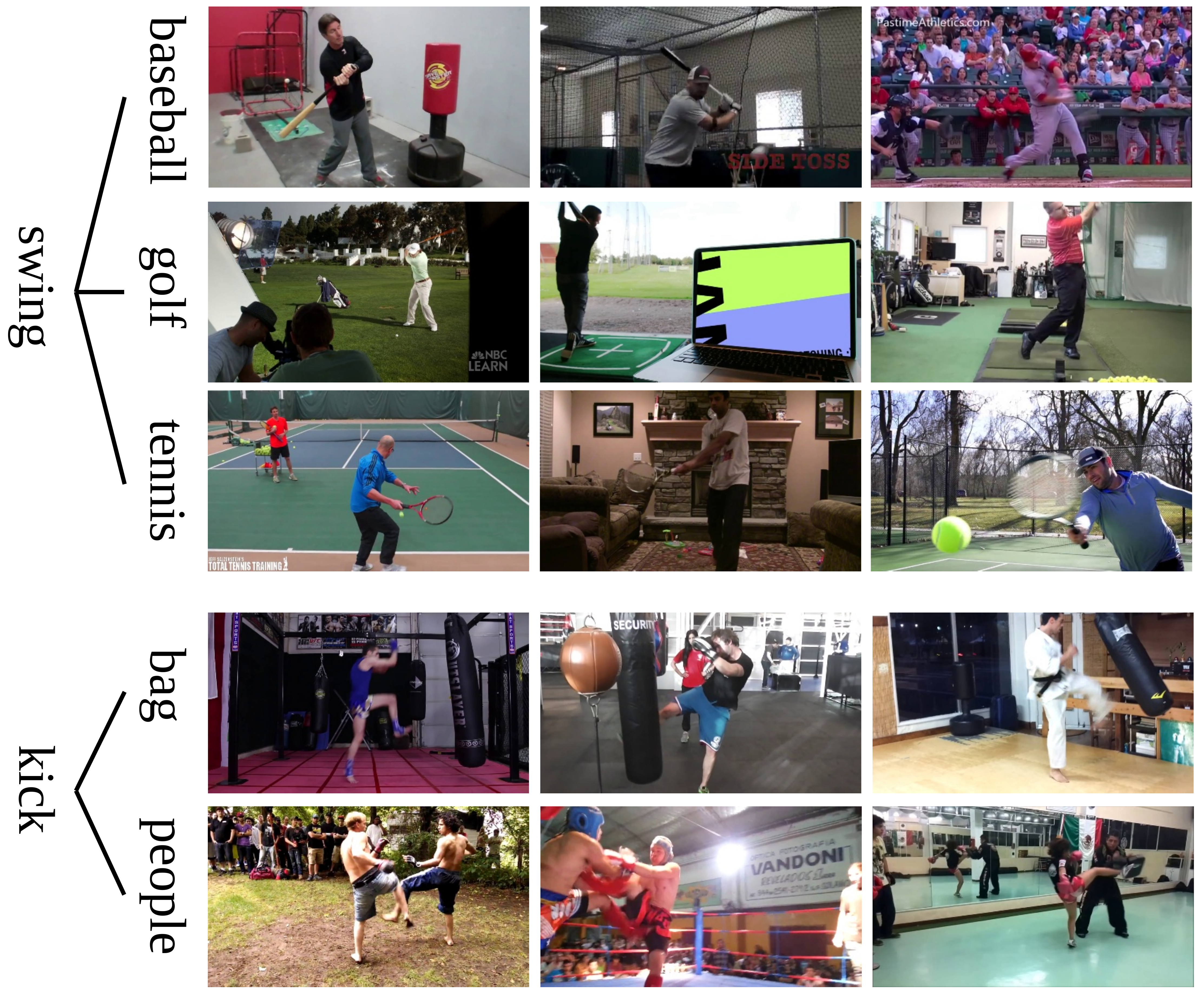}
\centering
\vspace{-0.1in}
\caption{\label{fig:dataset}
\footnotesize{Samples in our ACT dataset. The action classes are arranged in a two-layer hierarchy. For each class, we collected hundreds of video clips with large diversities.}}
\vspace{-0.25in}
\end{figure}

\textbf{Dataset collection.} To collect our dataset we used YouTube videos. We used 50 keywords to retrieve videos which belong to one of the 43 classes. For each keyword, we downloaded around 500 high quality videos which have length within 15 minutes. The videos were labeled by a commercial crowd-sourcing organization. We asked the workers to label the starting and ending frames for actions in the video. For each action class, we provided a detailed description and 3 annotation examples from different videos. To increase the diversity of actions, we required the workers to label no more than 8 action clips in each video, and between each clips there should be a temporal gap of at least 40 frames. We also set  temporal length limitations that each annotated clip should have at least 1 second and at most 10 seconds.  As Figure~\ref{fig:dataset} illustrates, our dataset has large intra-class diversities.

\textbf{Task design.} We design two tasks for our ACT dataset. The first task is standard action classification over 43 categories. The split used for this task included $65\%$ of videos as training and the rest as testing data, resulting in 7260 training videos and 3974 for testing in total.
The second proposed task for this dataset is cross-category generalization. For each of the 16 super-classes, we consider one of its sub-category as testing and the other sub-categories are used for training. For example, for super-class ``swinging'', we want to see if the model trained on swinging baseball and swinging golf can recognize swinging tennis as ``swinging''. We create 3 different random splits for the second task. There are around 7000 training samples and 4000 testing samples on average. Our dataset can be downloaded from the project website\footnote{\url{http://www.cs.cmu.edu/~xiaolonw/actioncvpr.html}}.
% ( \url{http://www.cs.cmu.edu/~xiaolonw/actioncvpr.html}).

% The selection for testing sub-categories is performed randomly.

\vspace{-2mm}
\section{Modeling Actions as Transformations}
\vspace{-1.5mm}

Given an input video $X$ consisting of $t$ frames, we denote each frame as $x_i$ and the whole video as $X =\{x_1,x_2,...,x_t\}$. We make an assumption that the precondition state of an action corresponds to the first $z_p$ frames and the effect of the action can be seen after from $z_e$ until the end. We denote precondition and effect frames as: $X_p = \{x_1\dots x_{z_p}\}$ and $X_e = \{x_{z_e}\dots x_{t}\}$. Note that we do not manually define how many frames are used of representing precondition and effect. Therefore $z_p$ and $z_e$ are treated as latent variables, which will be inferred automatically by our model during training and testing.

\begin{figure*}
\center
\includegraphics[width=0.9\textwidth]{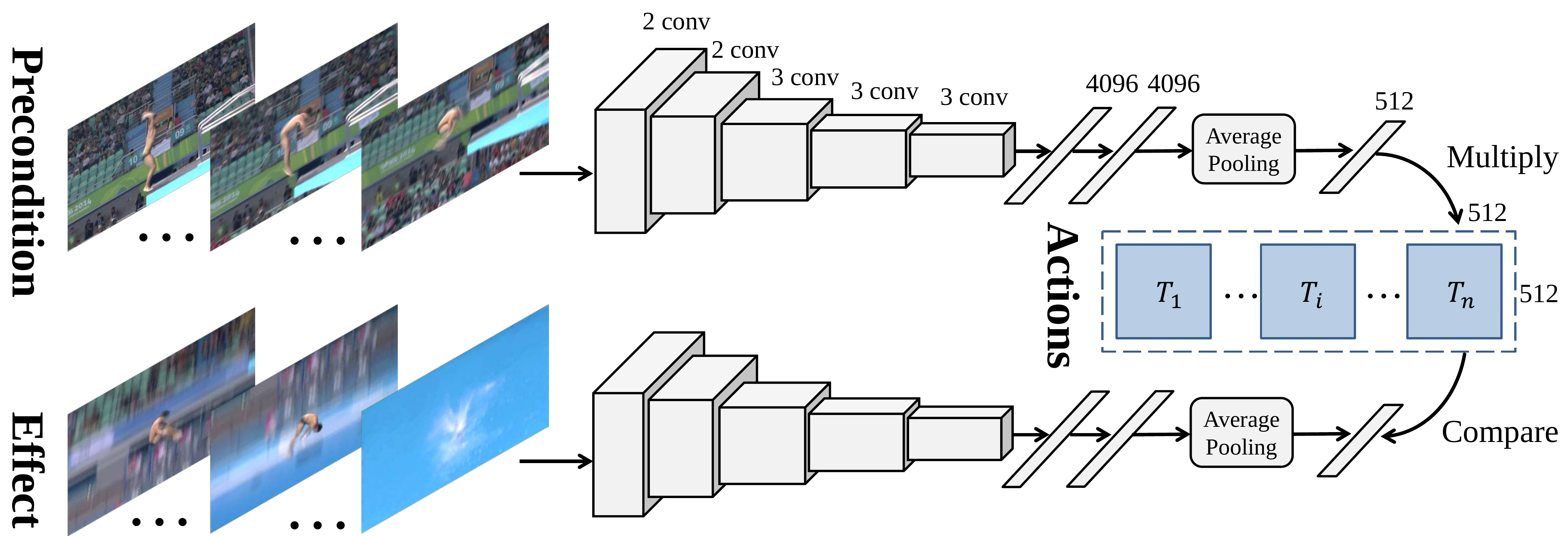}
\vspace{-0.11in}
\caption{\label{fig:network}
\footnotesize{Siamese network architecture. Given a video, we feed the precondition state frames to the top network and effect state frames to the bottom network. Each tower of the ConvNet computes the feature for each frame independently, and aggregates the features via average pooling. The pooling results are fully connected to 512-D embedding outputs. We apply $n$ transformations (actions) on the precondition embedding and compare with the effect embedding to decide the action class. } }
\vspace{-0.22in}
\end{figure*}

Instead of representing the precondition and effect by pixels and modeling the transformation in pixel space, we want to represent them using higher-level semantic features (e.g., the last fully connected layer of a ConvNet). The action then corresponds to transformation in  higher-level feature subspace. This can be modeled via a Siamese ConvNet as shown in Figure~\ref{fig:network}. Given the video frames for the precondition and effect states, we feed the frames of precondition state $X_p$ as inputs for the ConvNet on the top and the frames of  effect state $X_e$ are fed to the ConvNet on the bottom. For each tower of ConvNet, it computes the features for each frame independently, and then the features are aggregated via average pooling. We add a final $d$-dimensional fully connected layer after average pooling which represent the feature space where precondition, effect and the transformation between the two are modeled.  Formally, we use $f_p(X_p)$ to represent the $d$-dimension embedding for the precondition state generated from the network on the top (in Figure~\ref{fig:network}) given input $X_p$. For the network on the bottom, we represent the embedding for the effect state as $f_e(X_e)$ with the same dimension $d$ given input $X_e$.

Finally, we model the action as the transformation between these two states. Specifically, we use a linear transformation matrix to model this. For a dataset with $n$ categories of actions, we have a set of $n$ corresponding transformation matrices $\{T_1,...,T_n\}$ to represent them. Each $T_i$ is a $d \times d$ dimensions matrix. At the training time, given an input video
$X$ that belongs to action category $i$, we first obtain the embedding for the precondition and effect states of  the action as $f_p(X_p)$ and $f_e(X_e)$. We then apply the transformation matrix $T_i$ on the embedding of the precondition state as $ T_if_p(X_p)$, which is also a $d$-dimension vector. The objective of learning is making the distance $D( T_if_p(X_p), f_e(X_e))$ between the two embeddings small. Note that while training, the gradients are back propagated throughout the Siamese networks; thus we learn both the embedding space and the transformation simultaneously.

\textbf{Network Architecture}
We applied the VGG-16 network architecture~\cite{Simonyan15} for both sides of our Siamese network.  The VGG-16 network is a 16-layer ConvNet with 13 convolutional layers and 3 fully connected layers. As we mentioned before, we perform forward propagation for each video frame independently. We extract the feature of the second-to-last fully connected layer for each frame, which is a $4096$-D vector. In each side of our model, the features are aggregated via average pooling, and we use a final fully connected layer on these pooling outputs. The dimension of the last embedding layer outputs is $d=512$. In our model, we do not share the model parameters between two ConvNets. Intuitively, we want to learn different semantic representations for precondition and effect of actions.

\textbf{Two Stream Siamese: RGB and Optical Flow as Inputs. }
For each input frame, we rescale it by keeping the aspect ratio and make the smaller side $256$ pixels. To represent the video frames, we follow the same strategy as~\cite{2stream14}, which represents the frames with RGB images as well as optical flow fields. We train two separate models for RGB and optical flow fields as inputs. For the model using RGB images as inputs, the input size is $224 \times 224 \times 3$. For the model using optical flow as inputs, we represent each frame by stacking 10 optical flow fields extracted from 10 consecutive frames in the video starting from the current one. The optical flow field can be represented by a 2-channel image including horizontal and vertical flow directions. Therefore, the input size is  $224 \times 224 \times 20$ as mentioned in~~\cite{2stream14}.

{\bf Implementation Details.} During training and testing, we re-sample the videos to be 25 frames in length ($t=25$) as~\cite{2stream14}. Note that the optical flow fields are still computed in the original video without re-sampling. We also constrain the latent variables $z_p$ and $z_e$,  which are the indexes for the end frame of the precondition state and start frame of the effect state such that: $z_p \in [\frac{1}{3}t, \frac{1}{2}t )$ and $z_e \in (\frac{1}{2}t, \frac{2}{3}t]$.  

% In our experiments, we noticed that the choices of these parameters has no significant effect on the performance.

\vspace{-1mm}
\subsection{Training}
\vspace{-2mm}
We now introduce the training procedure for our model. Suppose we have a dataset of $N$ samples with $n$ categories $\{(X_i,y_i)\}_{i=1}^{N}$, where $y \in \{1,...,n\}$ is the label. Our goal is to optimize the parameters for the ConvNets and transformation matrices $\{T_i\}_{i=1}^n$. We also need to calculate the latent variables $z_p$ and $z_e$ for each sample. Thus, we propose an EM-type algorithm, which performs a two-step procedure in each iteration: (i) learning model parameters and (ii) estimating latent variables .

\textbf{(i) Learning model parameters.} We first discuss the optimization of model parameters given the latent variables. Our objective is to minimize the distance between the embedding of the precondition state after transformation and the embedding of the effect state computed by the second ConvNet. This can be written as:

\vspace{-0.2in}
{\small
\begin{eqnarray}\label{eq:cos_loss}
\min D( T_y f_p(X_p), f_e(X_e)),
\end{eqnarray}
}
where $T_y$ is the transformation matrix for the ground truth class $y$, $f_p(X_p)$ is the embedding for the precondition of the action and $f_e(X_e)$ is the embedding for the effect of the action. We use cosine distance here such that $D(v_1,v_2) = 1 - \frac{v_1 \cdot v_2} {\|v_1\| \|v_2\| }$ between any two vectors $v_1, v_2$.

To make our model discriminative, it is not enough to minimize the distance between two embeddings given the corresponding transformation $T_y$. We also need to maximize the distances for other incorrect transformations $T_i (i \neq y)$, which equals to minimize the negative of them. Thus, we have another term in the objective,

\vspace{-0.2in}
{\small
\begin{eqnarray}\label{eq:cos_loss2}
\min \sum_{i \neq y}^n max(0, M - D(T_if_p(X_p),f_e(X_e))),
\end{eqnarray}
}
where $M$ is the margin so that we will not penalize the loss if the distance is already larger than $M$. In our experiment, we set $M = 0.5$.
By combining Eq.(\ref{eq:cos_loss}) and Eq.(\ref{eq:cos_loss2}), we have the final loss for training. It is a contrastive loss given sample $(X,y)$ and latent variables $z_p$ and $z_e$ as inputs. We train our model with back-propagation using this loss.

\textbf{(ii) Estimating latent variables.} Given the model parameters, we want to estimate the end frame index $z_p$ of the precondition state and start frame index $z_e$ of the effect state. That is, we want to estimate the latent variables to give a reasonable temporal segmentation for the input video. Since the latent variable only depends on the ground truth action category; we only use the first term in the loss function, Eq.(\ref{eq:cos_loss}), to estimate $z_p$ and $z_e$ as follows:

\vspace{-0.2in}
{\small
\begin{eqnarray}\label{eq:cos_loss4}
(z_p^{\ast}, z_e^{\ast}) = argmin_{(z_p,z_e)} D( T_y f_p(X_p), f_e(X_e)),
\end{eqnarray}
}
where $(z_p^{\ast}, z_e^{\ast})$ are the estimation results given current model parameters. To estimate these latent variables, we use a brute force search through the space of latent variables. For computation efficiency, we first compute features for all frames, and then a brute force search only requires the average  pooling step to be redone for each configuration of $(z_p,z_e)$

\begin{small}
\begin{algorithm}[t]
\caption{Learning}
\label{alg:Framwork}
\begin{algorithmic}\footnotesize
\REQUIRE ~~\\
    $N$ videos with $n$ classes $\{(X_i,y_i)\}_{i=1}^{N}$, iteration number $Iter$.
\ENSURE ~~\\
    Parameters of ConvNets and transformation matrices.

\vspace{0.15cm}
\hspace{-0.4cm} Model initialization, $i = 0$.
\MYWHILE
    \STATE
    \begin{itemize}
\setlength{\itemsep}{1pt}
 \setlength{\parskip}{0pt}
 \setlength{\parsep}{10pt}
      \item[1.] Forward propagation and feature computation for each frame.
      \item[2.] Search latent variables with Eq.(\ref{eq:cos_loss4}).
      \item[3.] Back-propagations with the joint loss using Eqs. (\ref{eq:cos_loss}) and (\ref{eq:cos_loss2}).
      \item[4.] $i = i + 1$.
    \end{itemize}
\MYENDWHILE {$i = Iter$}

\end{algorithmic}

% \vspace{-0.1in}
\end{algorithm}

\end{small}

% pretrain
\textbf{Model pre-training.} We first initialize the ConvNets using ImageNet~\cite{ILSVRC15} pre-training and adapt them to action classification with fine-tuning as~\cite{Deep2stream15}. We transfer the convolutional parameters to both ConvNet towers and randomly initialize the fully connected layers in our model. We summarize the  learning procedure as Algorithm \ref{alg:Framwork}.

\textbf{Training detail discussions.} During training, we have not explicitly enforced the representations of $f_p(X_p)$ and $f_e(X_e)$ to be different. We have tried to use Softmax loss to classify precondition and effect as two different classes. We found it does not help improve the performance. The reason is that the two towers of networks are learned on different data with different parameters (no sharing) and initialization, which automatically leads to different representations. 

%Moreover, we also want to emphasize that the distance metric is learned via the deep networks since the parameters of the towers and the transformation matrices are learned end-to-end simultaneously. 

\vspace{-0.03in}
\subsection{Inference}
\begin{figure}
\includegraphics[width=0.5\textwidth]{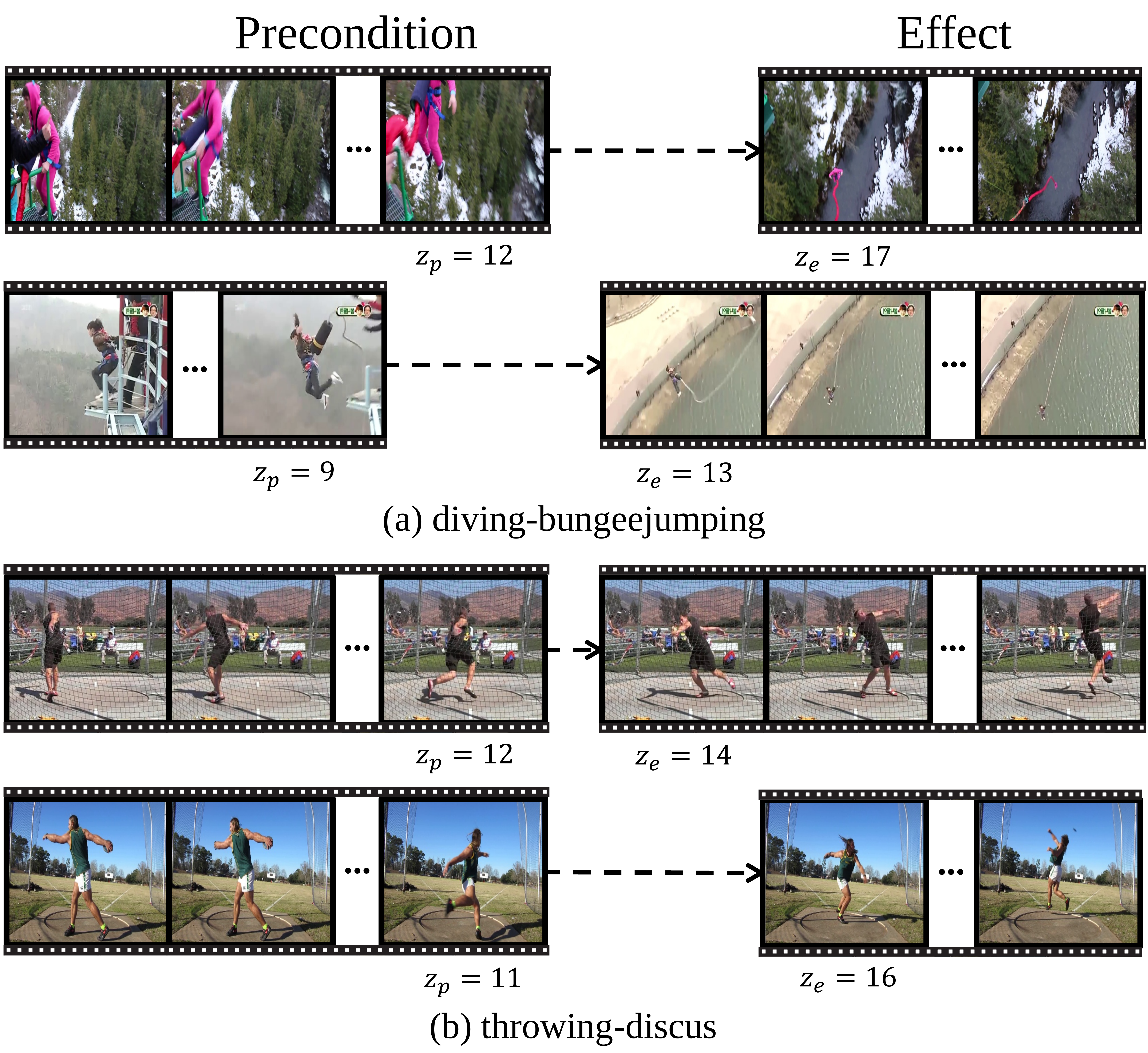}
\centering
\vspace{-0.25in}
\caption{\label{fig:latent} \footnotesize{
Temporal segmentation results after inference. $z_p$ and $z_e$ are the latent variables indexing the end frame for precondition and start frame for effect. Our model can provide reasonable temporal alignment of states between different videos in the same category. } }
\vspace{-0.2in}
\end{figure}

\vspace{-0.05in}
During inference, given a video $X$ and our trained model, our goal is to infer its action class label $y$ and segment the action into the precondition and effect states at the same time. The  inference  objective can be represented as,
\begin{eqnarray}\label{eq:cos_infer}
\min_{y,z_p,z_e} D( T_y f_p(X_p), f_e(X_e)).
\end{eqnarray}
More specifically, we first calculate the ConvNet feature before the average pooling layer for all the frames. Note that the first half of the frames are fed into the network for the precondition state and the second half of the frames are fed into the network for the effect state. We do a brute force search over the space of $(y,z_p,z_e)$ to estimate the action category and segmentation into precondition and effect. We visualize reasonable segmentation results for precondition and effect states during inference as Figure ~\ref{fig:latent}.

\textbf{Model fusion.} We perform the inference with the models using RGB images and optical flow as inputs separately. For each video, we have $n$ distance scores from each model. We fuse these two sets of scores by using weighted average. As suggested by~\cite{Deep2stream15} we weight the flow twice as much as the RGB ConvNet.

\vspace{-1.5mm}
\section{Experiment}
\vspace{-1mm}

\vspace{-1mm}
In this section, we first introduce the details of the datasets and our experimental settings. Then we provide quantitative and qualitative results on different datasets.

\textbf{Datasets.} We evaluate our method on three datasets, including UCF101~\cite{UCF12}, HMDB51~\cite{HMDB11} and  our ACT  dataset. UCF101 dataset contains 13320 videos with 101 classes. The HMDB51 dataset is composed of 6766 videos with 51 action categories. For both datasets, we follow the standard evaluation criteria using 3 official training/testing splits. Our ACT dataset has 11234 video clips from 43 action classes and 16 super-classes. For this dataset, we conduct our experiments on two tasks. The first task is standard action classification of 43 categories. The second task is to test the generalization ability of our model. We conduct the experiments using 3 training/testing splits for 16-class classification. For each super-class, one sub-category is used for testing and the others for training.

\textbf{Implementation Details.} We first pre-train our networks using Softmax loss on action classification as the Two Stream baseline~\cite{Deep2stream15} and then transfer the convolutional parameters to our model. For HMDB51 dataset, as the number of training videos are relatively small (around 3.6K), we borrowed the ConvNet trained on UCF101 dataset with Softmax loss to initialize our model as ~\cite{TDD15,2stream14}. For the ACT dataset, we did not use UCF101 or HMDB51 during pre-training. Note that there are two types of inputs for our model: RGB and optical flow. We trained two different models for different inputs.  For the model using RGB images as input, we set the initial learning rate as $10^{-5}$ and reduce it by 10 times
after every $10K$ iterations and stop at $30K$ iterations. For the model using optical flow, the initial learning rate is set to $10^{-5}$. We reduce learning rate by order of 10 for every $20K$ iterations and stop at $50K$ iterations. We set the size of the batch to $50$ videos.
To compute the optical flow, we apply the TVL1 optical flow algorithm~\cite{Zack07} and discretize the values to the range of $[0,255]$.

\vspace{-1mm}
\subsection{Experimental Results.}
\vspace{-1mm}

\begin{table}
\begin{center}
\resizebox{0.5\textwidth}{!}{
\begin{tabular}{|c|c|c|c|}
  \hline
  Method & RGB & Optical Flow & Fusion\\ \hline
  Two Stream ~\cite{2stream14} & 73.0\% &  83.7\% &  88.0\%  \\
  Two Stream (VGG16)~\cite{Deep2stream15} & 78.4\% &  87.0\% &  91.4\%  \\
  Ours      & \textbf{80.8\%} & \textbf{87.8\%} & \textbf{92.4\%}  \\
  \hline
\end{tabular}
}

\vspace{-2mm}
\caption{ \footnotesize{Average accuracies on UCF101 over 3 splits.} }
\label{tbl:result_ucf}
\end{center}
\end{table}

\begin{table}

\vspace{-5mm}
\begin{center}
\resizebox{0.5\textwidth}{!}{
\begin{tabular}{|c|c|c|c|}
  \hline
  Method (for split1) & RGB & Optical Flow & Fusion\\ \hline
  Two Stream (VGG16)~\cite{Deep2stream15} & 79.8\% &  85.7\% &  90.9\%  \\ \hline
  Two Stream First Half & 80.0\% &  84.7\% & \multirow{2}{*}{90.0\%}   \\
  Two Stream Second Half & 79.5\% & 84.8\% &          \\ \hline
  Ours      & \textbf{81.9\%} & \textbf{86.4\%} & \textbf{92.0\%}  \\
  \hline
\end{tabular}
}

\vspace{-2mm}
\caption{ \footnotesize{Comparing method training different networks for different parts of the videos on UCF split1.} }
\label{tbl:result_ucf1}
\end{center}
\vspace{-7mm}
\end{table}

\textbf{UCF101 dataset.} The accuracies over 3 training/testing splits of UCF101 are shown in Table~\ref{tbl:result_ucf}. The Two Stream~\cite{2stream14} method using an 8-layer CNNs achieves $88.0\%$. By replacing the base network with VGG-16 architecture, ~\cite{Deep2stream15} obtains $91.4\%$ accuracy. We use~\cite{Deep2stream15} as a baseline for comparisons. Our model outperforms both models using RGB images and optical flow fields. After fusion of two models, we achieve state of the art performance at $92.4\%$. See Table~\ref{tbl:stoa} for more comparisons.

To further analyze our results and factor out the contributions of the appearance of the precondition and effect segments, we perform an ablative analysis on the first split of UCF101. We train different networks using Softmax loss for different segments of the videos independently. For each video, we cut it by half and train one network using the first half of frames and another network is trained using the second half. In total we train 2 RGB ConvNets and 2 optical flow  ConvNets. For model fusion, we average the results from these 4 networks. We show the results in Table~\ref{tbl:result_ucf1}. Compared to the baseline method~\cite{Deep2stream15}, we find that the performance decreases most of the time and the fusion model is $0.9\%$ worse than the Two Stream baseline. This is mainly because decreasing the number of training samples leads to over-fitting. This analysis shows that modeling the precondition and effect separately does not work as well as the baseline, but modeling the transformations between them is $1.1\%$ better than the Two Stream baseline.

\begin{table}
\begin{center}
\resizebox{0.5\textwidth}{!}{
\begin{tabular}{|c|c|c|c|}
  \hline
  Method & RGB & Optical Flow & Ave Fusion\\ \hline
  Two Stream~\cite{2stream14}  & 40.5\% &  54.6\% &  58.0\%  \\
  Two Stream (VGG16)& 42.2\% &  55.0\% &  58.5\%  \\
  Ours      & \textbf{44.1\%} & \textbf{57.1\%} & \textbf{62.0\%}  \\
  \hline
\end{tabular}
}

\vspace{-2mm}
\caption{ \footnotesize{Average accuracies on HMDB51 over 3 splits. } }
\label{tbl:result_hmdb}
\end{center}
\vspace{-6.5mm}
\end{table}

\textbf{HMDB51 dataset.} For HMDB51, we also report the average accuracies over 3 splits in Table~\ref{tbl:result_hmdb}. As a baseline, we apply Two Stream method~\cite{2stream14,Deep2stream15} with VGG-16. The baseline method is better than \cite{2stream14} with average fusion. However, it is not as good as the best results ($59.4\%$) in~\cite{2stream14} using SVM for model fusion.  We show that our method has $1.9\%$ gain for models using RGB images and $2.1\%$ gain for model using optical flow. After model fusion, our method reach $62\%$ accuracy, which is $3.5\%$ better than the baseline.

More interestingly, we show that our method and the Two Stream method are complimentary to each other by combining the classification results from two methods. Since our outputs are distances and the Two Stream outputs are probabilities, we convert the distances outputs to probabilities. To do this, we extract the precondition embedding after ground truth transformation and the effect embedding on training data. We concatenate the two embeddings and train a Softmax classifier. In testing, we apply the same inference method as before and use the newly trained Softmax classifier to generate the probabilities, which gives the same performance as before. We  average the outputs from two methods with equal weights as our final result. As Table~\ref{tbl:stoa} shows, by combining our method and the Two Stream method we have $1.4\%$ boost, which leads to $63.4\%$ accuracy. However, the state of the art results in this dataset are based on Improved Dense Trajectories (IDT) and Fisher vector encoding ($63.7\%$~\cite{Basura15} and $66.8\%$~\cite{Peng14}).

%Note that we use the same initialization for the convolution layers in our model and the Two Stream model. 

\begin{table}
\small
\centering
\begin{tabular}{|lr|lr|}
\hline
\multicolumn{2}{|c|}{HMDB51} & \multicolumn{2}{|c|}{UCF101} \\
\hline
\hline
STIP+BoW \cite{HMDB11} & 23.0\% & STIP+BoW \cite{UCF12} & 43.9\% \\
\hline
DCS+VLAD \cite{Jain13} & 52.1\% & CNN\cite{Karpathy14} & 65.4\% \\
\hline
VLAD Encoding \cite{Wu14} & 56.4\% &  IDT+FV \cite{WangIDT13} & 85.9\% \\
\hline
IDT+FV \cite{WangIDT13} & 57.2\% & IDT+HSV \cite{iDTHSV14} & 87.9\% \\
\hline
Two Stream \cite{2stream14} & 59.4\%   & Two Stream \cite{2stream14} & 88.0\% \\
\hline
IDT+HSV \cite{iDTHSV14} & 61.1\% & LSTM with  & 88.6\%  \\
&&Two Stream~\cite{LSTMgoogle15}& \\
\hline
TDD+FV \cite{TDD15} & 63.2\% &  TDD+FV \cite{TDD15} & 90.3\%  \\
\hline
Two Stream & 58.5\% & Hybrid LSTM~\cite{Wu15} & 91.3\% \\
 (VGG16) by us& &  & \\
\hline
Ours & 62.0\%  & Two Stream   & 91.4\%  \\
&& (VGG16)~\cite{Deep2stream15} & \\
\hline
Ours+Two Stream & \textbf{63.4\%} & Ours & \textbf{92.4\%} \\
\hline
\end{tabular}
\vspace{-2.5mm}
\caption{\footnotesize{Comparison to state of the art results. } }
\label{tbl:stoa}
\vspace{-0.5mm}
\end{table}

\begin{table}
\vspace{-1mm}
\begin{center}
\resizebox{0.45\textwidth}{!}{
\begin{tabular}{|c|c|c|c|}
  \hline
  Method & RGB & Optical Flow & Fusion\\ \hline
  Two Stream  & 66.8\% &  71.4\% &  78.7\%  \\
  LSTM+Two Stream & 68.7\% &  72.1\% &  78.6\%  \\
  Ours      & \textbf{69.5\%} & \textbf{73.7\%} & \textbf{80.6\%}  \\
  \hline
\end{tabular}
}

\vspace{-2mm}
\caption{\footnotesize{Accuracies for the first task on ACT dataset.}}
\label{tbl:result_act}
\end{center}
\vspace{-8mm}
\end{table}

\begin{table*}
\begin{center}
\resizebox{1\textwidth}{!}{
\begin{tabular}{|c|c|c|c|c|c|c|c|c|c|c|c|c|}
  \hline
  &\multicolumn{4}{|c|}{RGB} & \multicolumn{4}{c|}{Optical Flow} &  \multicolumn{4}{c|}{Fusion} \\ \hline
  Model & Split1 & Split2 & Split3 & Average & Split1 & Split2 & Split3 & Average & Split1 & Split2 & Split3 & Average \\ \hline
  Two Stream & 48.3\% & 53.2\% & 49.7\% & 50.4\% & 53.7\% & 56.6\% & 56.3\% &  55.5\% & 59.5\% & 67.6\% & 62.2\% &  63.2\% \\
  LSTM+Two Stream & 47.8\% & 53.6\% & 50.2\% & 50.5\% & 55.6\% & 57.0\% & 57.4\% & 56.7\% & 59.9\% & 67.3\% & 61.1\% & 62.8\%\\
  Ours & 49.3\% & 54.8\% & 52.5\% & {\bf 52.2\%} & 57.6\% & 59.5\% & 60.4\% & {\bf 59.2\%} & 62.4\% & 68.7\% & 65.4\% & {\bf 65.5\%} \\
  \hline
\end{tabular}
}
\vspace{-2.5mm}
\caption{\footnotesize{Accuracies over 3 splits for the second task on ACT dataset. } }
\label{tbl:result_act2}
\end{center}
\vspace{-6mm}
\end{table*}

\textbf{ACT dataset.} We evaluate on two tasks in this dataset as proposed in the Dataset section. Motivated by the recent literature~\cite{LSTMgoogle15,Wu15,Donahue15} in using LSTM to model the temporal information of videos, we also conduct experiments using the LSTM as another baseline for comparison. All the ConvNets are based on the VGG-16 architecture. For this LSTM approach, we first perform forward propagation with Two Stream model on each of the 25 frames and extract the feature from the fully connected layer before the classification outputs. The features are fed into LSTM to generate the classification scores. During training, we train the LSTM and Two Stream models jointly.

The first task is the standard action classification over 43 classes. As Table~\ref{tbl:result_act} illustrates, the LSTM method is better than the Two Stream method given RGB and optical flow inputs individually, however the results after fusion are very close. On the other hand, our method reach $80.6\%$, $1.9\%$ higher than the baseline.

For the second task, we examine the cross category generalization ability of our model. We perform 16-class classification on 3 different splits. For each class, one sub-category is left out for testing and the other sub-categories are used for training. As Table~\ref{tbl:result_act2} shows, compared to the Two Stream baseline, we have $1.9\%$ improvements in models using RGB images as inputs and $3.7\%$ improvements using optical flow fields as inputs in average. After fusing the two models with different inputs, we have $65.5\%$ accuracy, $2.3\%$ higher than the baseline. From these results, we can see that modeling actions as transformations leads to better category generalization. To explain why our model works better, we perform several visualizations in the following section.

\begin{figure}
\includegraphics[width=0.45\textwidth]{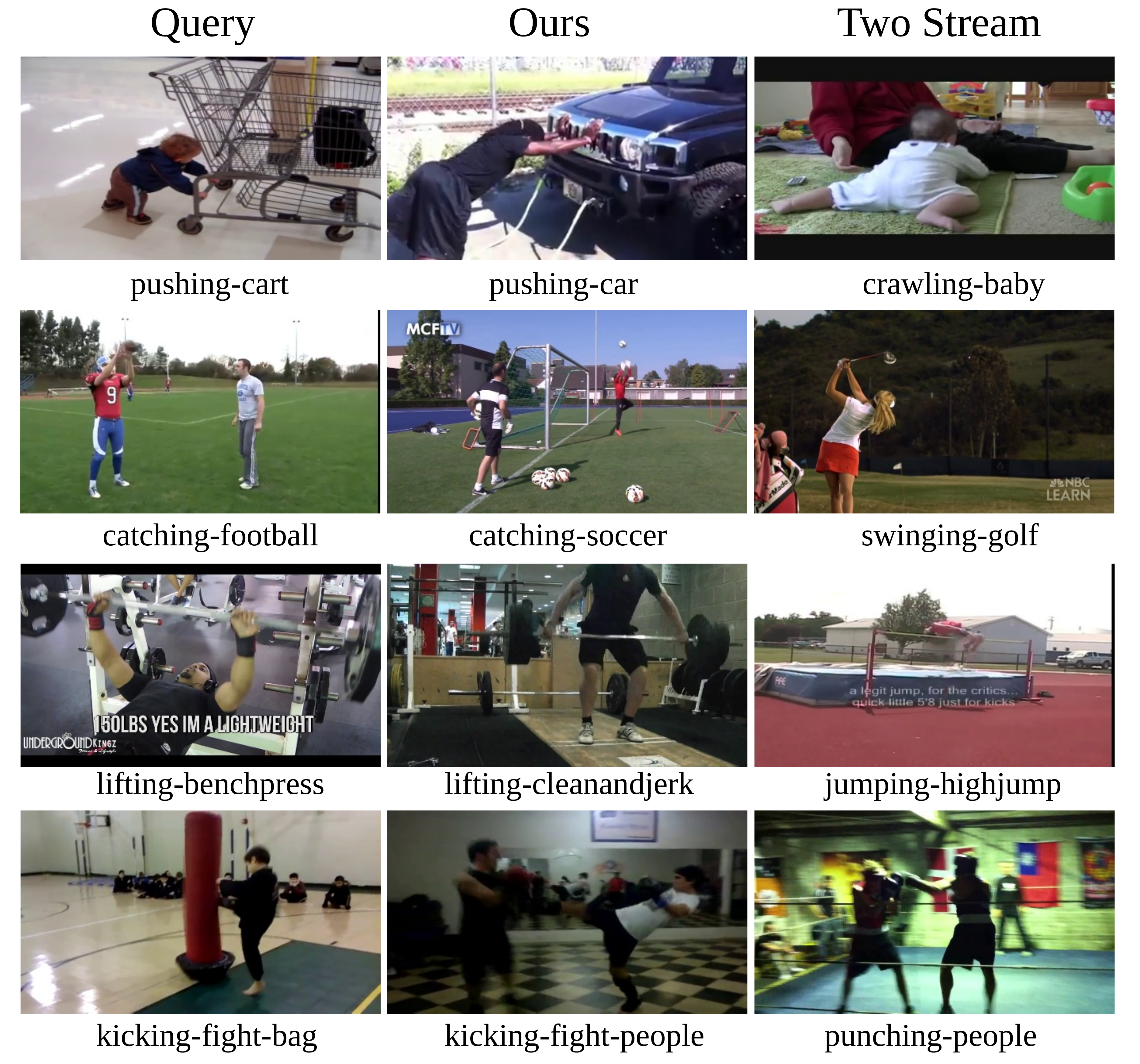}
\centering
\vspace{-0.1in}
\caption{\label{fig:confuse} \footnotesize{
Nearest neighbor test. Given the query videos on the left, our method can retrieve semantically related videos in the middle while the Two Stream method retrieves videos with similar appearance and motion on the right.} }
\vspace{-0.2in}
\end{figure}

\subsection{Visualization}
\vspace{-1.5mm}

\begin{figure*}
\center
\includegraphics[width=0.85\textwidth]{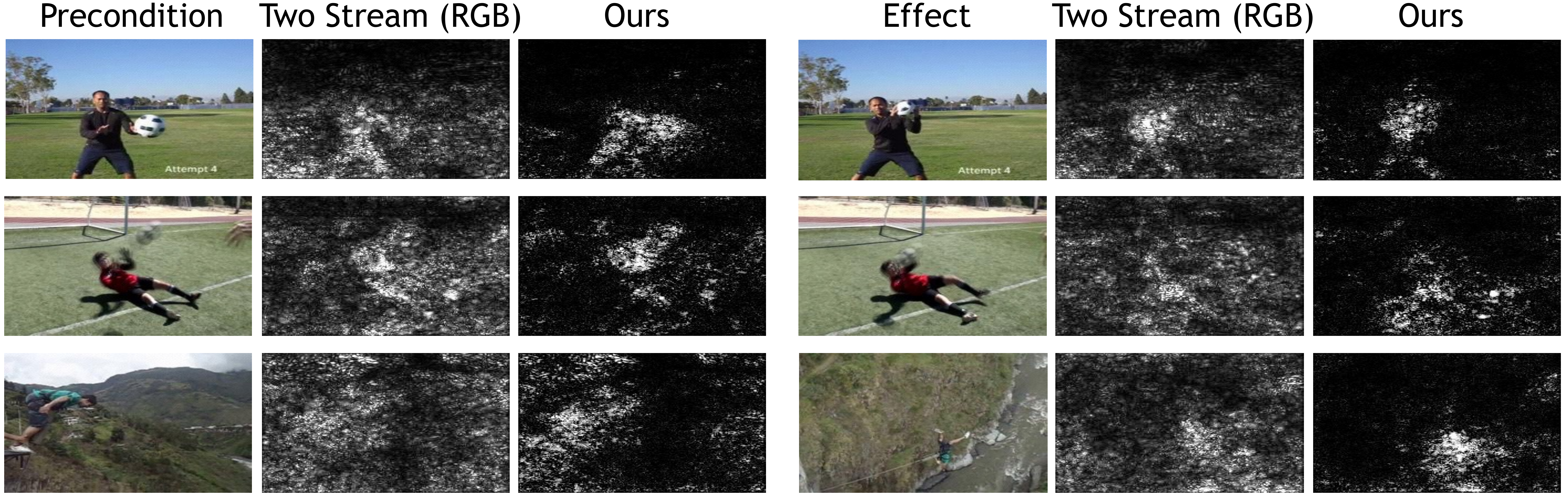}
\vspace{-0.1in}
\caption{\label{fig:saliency} \footnotesize{
Given the precondition and effect frames as inputs, we visualize the magnitude of gradients via back-propagation. While the network trained with RGB images in Two Stream method are sensitive to the object and scenes, our model focus more on the changes.} }
\vspace{-0.1in}
\end{figure*}

\begin{figure*}
\center
\includegraphics[width=0.85\textwidth]{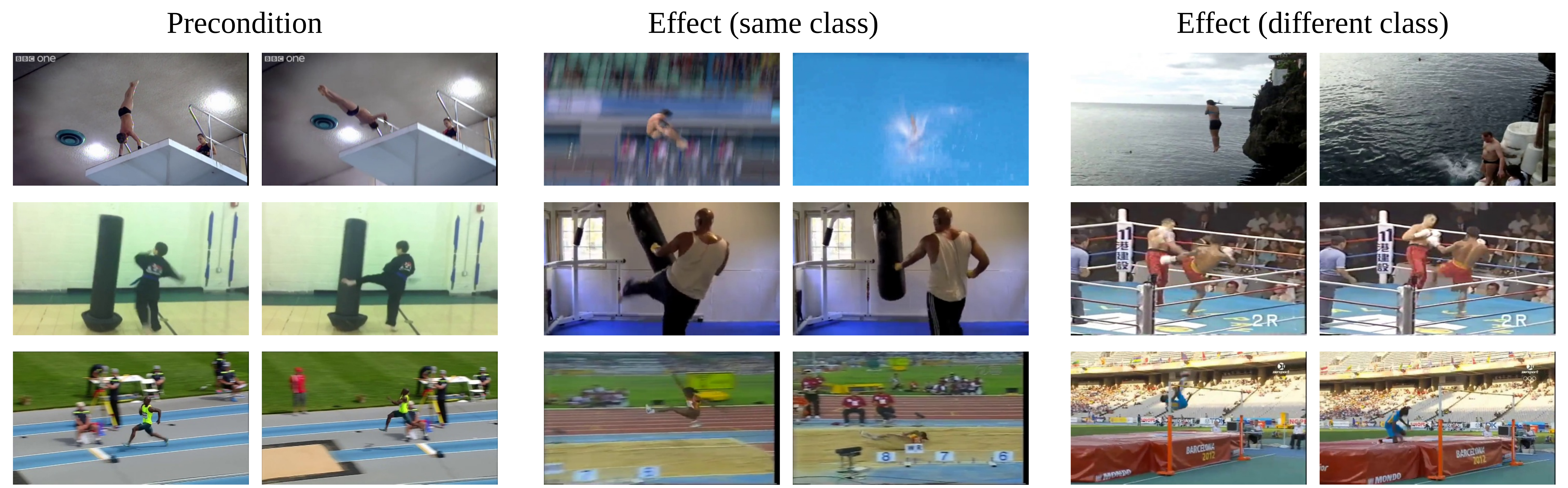}
\vspace{-0.1in}
\caption{\label{fig:prediction} \footnotesize{ Prediction results. Given the precondition frames of test videos on the left, we retrieve the effect frames of training samples. In the middle are the retrieval results with the same class as the query, on the right are results from different classes.} }
\vspace{-0.2in}
\end{figure*}

Just showing quantitative results does not give a full understanding of our method. What is our model learning? In which way our model is doing better? To answer these questions, we perform several visualization tasks on our ACT dataset.

\textbf{Nearest neighbor test.} In this experiment, we want to visualize  in what cases our method is doing better via nearest neighbor on the features, which are extracted by the models trained in the second task of the ACT dataset.
To extract the feature with our model, we perform inference on the video and concatenate the precondition embedding after transformation with the effect embedding. For the baseline Two Stream method, we extract the second-to-last fully connected layer feature and compute the average feature over the frames in the video. For both methods, we extract the feature for the RGB and optical flow models and then average them. As shown in Figure~\ref{fig:confuse}, given the query from testing samples on the left, we show the retrieved videos from the training videos. In the middle is our result and the retrieval results of the Two Stream method are on the right. For instance, on the first row, the query is a baby pushing a cart, we can retrieve the video in which a man is pushing a car, while the Two Stream method gets the baby crawling on the floor as the nearest neighbor. As the Two Stream method tries to match videos with similar appearance and motions, modeling the actions as transformations give us more semantically related retrieval results.

\textbf{Visualization of what the models focus on.}
In this experiment, we visualize what the models focus on during inference using similar techniques as~\cite{Simonyan13,Gan15}. We select one frame from the precondition state and another frame from the effect state, we feed these two RGB images into two towers of our model. We calculate the transformation loss given the video label and perform back-propagation.
We visualize the magnitude of the gradients passed to the input image. This visualization shows which parts of the image force the network to decide on an action class. We visualize the RGB network from the Two Stream method (using Softmax loss) in the same way. As Figure~\ref{fig:saliency} shows, given the precondition and effect frames, our model focuses more on the changes happening to the human while the RGB network from the Two Stream method performs classification based on the entire scene. In the first row, our model focus on the man and the ball  when the ball is flying towards to him. As the man catches the ball, our model has high activations on the man with the ball. In the second row, our model can also capture locations where the ball is flying towards the man. In both cases, the direct classification model fires on the whole scene. 
%Interestingly, our model notices the change of the feet and the shadow. 
%because the green field gives the model a strong cue of what the people is doing.

\textbf{Visual prediction.} Given the precondition embedding after transformation on the testing video, we retrieve the effect embedding from the same or different classes among the training data. We use the model trained on the first task of the experiment on the ACT dataset (43-class classification). If our model is actually learning the transformation, then it should be able to find effects frames that are similar to what we expect from applying the transformation to the precondition frames. We visualize the results in Figure~\ref{fig:prediction}. In each row, the first two images are the beginning and ending frames of precondition segments of  testing videos. The two images in the middle are the retrieval results with the same label as query. The two images on the right are the retrieval results with different class label as query. We show that our method can retrieve   reasonable and smooth prediction results. For example, when our model is asked to predict what is going to happen when a man jumps from a diving board, it retrieves getting into the pool as the within-category result, and getting into the lake as the  cross-category result. This is another information showing that our model is in fact learning actions as transformations from preconditions to effects.

\vspace{-2.5mm}
\section{Conclusion}
\vspace{-2mm}
We propose a novel representation for action as the transformation from precondition to effect.  We show promising action recognition results on UCF101, HMDB51 and ACT datasets. We also show that our model has better ability in cross category generalization. We show qualitative results which indicate that our method can explicitly model an action as the  change or transformation it brings to the environment.

{\footnotesize
\noindent {\bf Acknowledgement}: This work was partially supported by ONR N000141310720, NSF IIS-1338054, ONR MURI N000141010934 and ONR MURI N000141612007. This work was also supported by Allen Distinguished Investigator Award and gift from Google. The authors would like to thank Yahoo! and Nvidia for the compute cluster and GPU donations respectively. The authors would also like to thank Datatang for labeling the ACT  dataset. 
}

\bibliographystyle{ieee}
\bibliography{local}

\end{document}